\documentclass{article}

\usepackage[nonatbib, preprint]{neurips_2019}
\usepackage[utf8]{inputenc} 
\usepackage[T1]{fontenc}    
\usepackage{url}            
\usepackage{booktabs}       
\usepackage{amsfonts}       
\usepackage{nicefrac}       
\usepackage{microtype}      
\usepackage{graphicx}
\usepackage{tcolorbox}
\usepackage{amsmath}
\usepackage{float}
\usepackage{subfig}
\usepackage{color}

\usepackage[style=numeric, sorting=none]{biblatex}

\usepackage{tikz}
\usetikzlibrary{shapes.geometric, arrows}
\tikzstyle{startstop} = [rectangle, rounded corners, minimum width=3cm, minimum height=1cm,text centered, draw=black, fill=red!30, text width = 3cm]
\tikzstyle{io} = [trapezium, trapezium left angle=70, trapezium right angle=110, minimum width=3cm, minimum height=1cm, text centered, draw=black, fill=blue!30]
\tikzstyle{process} = [rectangle, minimum width=3cm, minimum height=1cm, text centered, draw=black, fill=orange!30, text width=4cm]
\tikzstyle{decision} = [diamond, minimum width=3cm, minimum height=1cm, text centered, draw=black, fill=green!30, text width = 2cm]
\tikzstyle{arrow} = [thick,->,>=stealth]

\title{Neural networks grown and self-organized by noise}

\author{%
  Guruprasad Raghavan
  \\
  Department of Bioengineering\\
  Caltech\\
  Pasadena, CA 91125\\
  \texttt{graghava@caltech.edu} \\
  \And
  Matt Thomson \\
  BBE \\
  Caltech \\
  Pasadena, CA 91125  \\
   \texttt{mthomson@caltech.edu} \\
}

\begin{document}

\maketitle

\begin{abstract}
Living neural networks emerge through a process of growth and self-organization that begins with a single cell and results in a brain, an organized and functional computational device. Artificial neural networks, however, rely on human-designed, hand-programmed architectures for their remarkable performance. Can we develop artificial computational devices that can grow and self-organize without human intervention? In this paper, we propose a biologically inspired developmental algorithm that can ‘grow’ a functional, layered neural network from a single initial cell. The algorithm organizes inter-layer connections to construct a convolutional pooling layer, a key constituent of convolutional neural networks (CNN’s).  Our approach is inspired by the mechanisms employed by the early visual system to wire the retina to the lateral geniculate nucleus (LGN), days before animals open their eyes.  The key ingredients for robust self-organization are an emergent spontaneous spatiotemporal activity wave in the first layer and a local learning rule in the second layer that ‘learns’ the underlying activity pattern in the first layer.  The algorithm is adaptable to a wide-range of input-layer geometries, robust to malfunctioning units in the first layer, and so can be used to successfully grow and self-organize pooling architectures of different pool-sizes and shapes. The algorithm provides a primitive procedure for constructing layered neural networks through growth and self-organization.   Broadly, our work shows that biologically inspired developmental algorithms can be applied to autonomously grow functional `brains' in-silico.

\end{abstract}

\begin{refsection}

\section{Introduction}

Living neural networks in the brain perform an array of computational and information processing tasks including sensory input processing \cite{glickfeld2017higher,peirce2015understanding}, storing and retrieving memory \cite{tan2017development,denny2017engrams}, decision making \cite{hanks2017perceptual, padoa2017orbitofrontal}, and more globally, generate the general phenomena of ``intelligence''. In addition to their information processing feats, brains are unique because they are computational devices that actually self-organize their intelligence. In fact brains ultimately grow from single cells during development. Engineering has yet to construct artificial computational systems that can self-organize their intelligence. In this paper, inspired by neural development, we ask how artificial computational devices might build themselves without human intervention. 

Deep neural networks are one of the most powerful paradigms in Artificial Intelligence. Deep neural networks have demonstrated human-like performance in tasks ranging from image and speech recognition to game-playing \cite{koch2015siamese, song2015end,silver2017mastering}. Although the layered architecture plays an important role in the success \cite{saxe2011random} of deep neural networks, the widely accepted state of art is to use a hand-programmed network architecture \cite{krizhevsky2012imagenet} or to tune multiple architectural parameters, both requiring significant  engineering investment. Convolutional neural networks, a specific class of DNNs, employ a hand programmed architecture that mimics the pooling topology of neural networks in the human visual system.

\begin{figure}[H]
 \vspace{-3mm}
    \centering
    \subfloat[]{{\includegraphics[width=5cm]{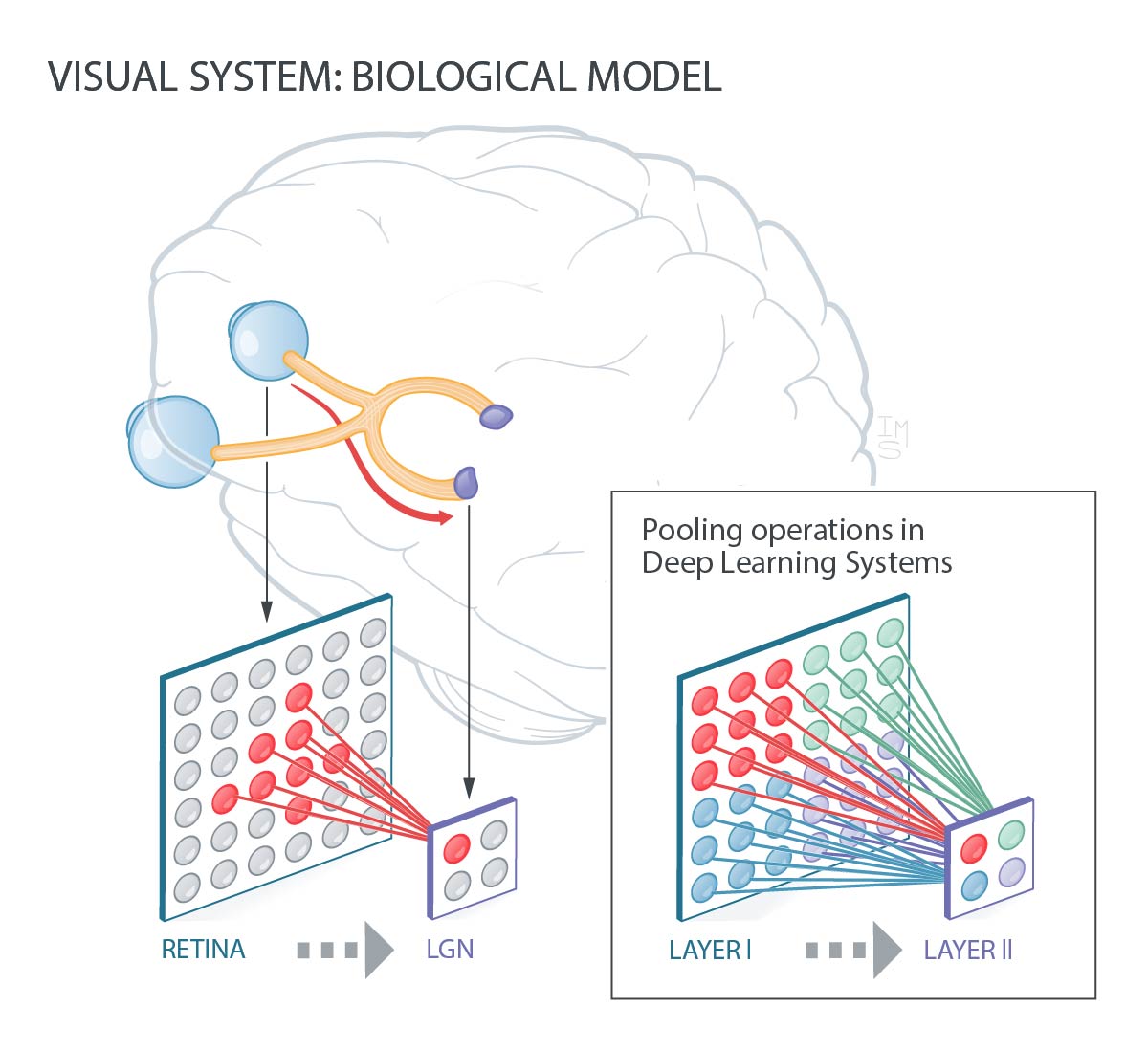}}}
    \qquad
    \subfloat[]{{\includegraphics[width=5cm]{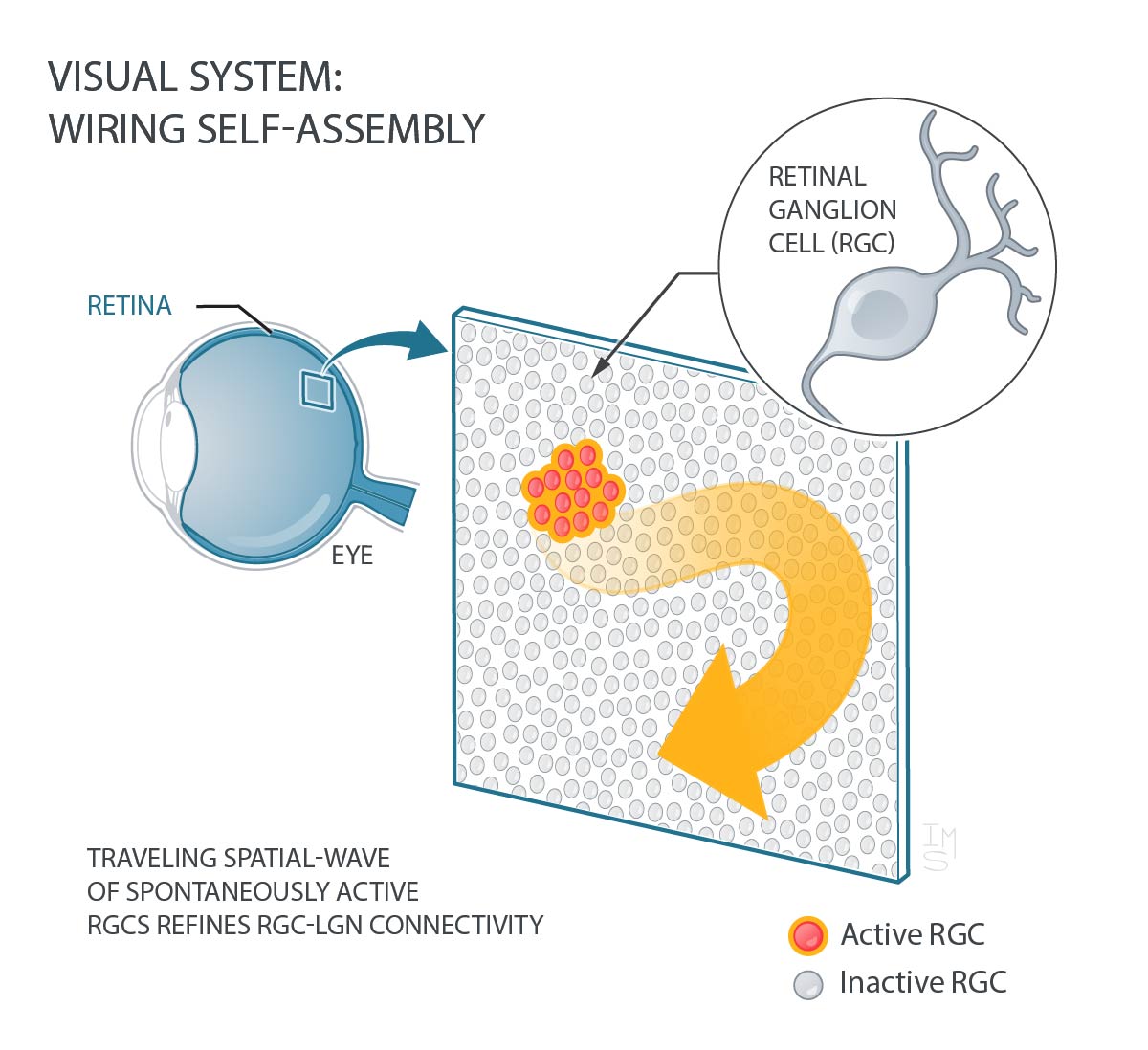}}}
    \caption{\textbf{Wiring of the visual circuitry} (a) Spatial pooling observed in wiring from the retina to LGN and in CNN's. (b) Synchronous Spontaneous bursts (retinal waves) in the light-insensitive retina serve as a signal for wiring retina to the brain.}
    \vspace{-3mm}
    \label{fig:bioInspired} 
\end{figure}

Here, we develop strategies for \textit{growing a neural network} autonomously from a single computational ``cell" followed by \textit{self-organization} of its architecture by implementing a wiring algorithm inspired by the development of the mammalian visual system. The visual circuity, specifically the wiring of the retina to the lateral geniculate nucleus (LGN) is stereotypic across organisms, as the architecture always enforces pooling (retinal ganglion cells (RGC's) pool their inputs to LGN cells) and retinotopy. The pooling architecture (figure-\ref{fig:bioInspired}a) is robustly established early in development through the emergence of spontaneous activity waves (figure-\ref{fig:bioInspired}b) that tile the light insensitive retina \cite{meister1991synchronous}. As the synaptic connectivity between the different layers in the visual system get tuned in an activity-dependent manner, the emergent activity waves serve as a signal to alter inter-layer connectivity much before the onset of vision. 

The main contribution of this paper is that we propose a developmental algorithm inspired by visual system development to \textit{grow and self-organize a pooling architecture}, a key feature of the convolutional neural network (CNN). Once a pooling architecture emerges, any non-linear function can be implemented by units in the second layer to morph it into functioning as a convolution or a max/average pooling. We show that our algorithm is adaptable to a wide-range of input-layer geometries, robust to malfunctioning units in the first layer and can grow pooling architectures of different shapes and sizes, making it capable of countering the key challenges accompanying growth. We also demonstrate that `grown' networks are functionally similar to that of hand-programmed pooling networks on conventional image classification tasks. As CNN's represent a model class of deep networks, we believe the developmental strategy can be broadly implemented for the self-organization of intelligent systems.

\section{Related Work}

Self-organization of neural networks dates back many years, with the first demonstration being Fukushima's neocognitron \cite{fukushima1988neocognitron,fukushima1991handwritten}, a hierarchical multi-layered neural network capable of visual pattern recognition through learning. Although weights connecting different layers were modified in an unsupervised fashion, the network architecture was hard-coded, inspired by Hubel and Wiesel's \cite{hubel1963shape} description of simple and complex cells in the visual cortex. This development inspired modern-day convolutional neural networks (CNN) \cite{lecun1990handwritten}. Although CNN's performed well on image-based tasks, they had a fixed, hand-designed architecture whose weights were altered by back-propagation. This changed with the advent of neural architecture search \cite{elsken2018neural}, as neural architectures became malleable to tuning by neuro-evolution strategies \cite{stanley2002evolving, real2017large, real2018regularized}, reinforcement learning \cite{zoph2016neural} and multi-objective searches \cite{elsken2018efficient, zhou2018neural}. These strategies have been successful in training networks that perform significantly much better on CIFAR-10, CIFAR-100 and Image-Net datasets. As the objective function being maximized is the predictive performance on these datasets the networks evolved may not generalize well to multiple datasets. On the contrary, biological neural networks in the brain grow architecture that can generalize very well to innumerable datasets. Neuroscientists have been very interested in how the architecture in the visual cortex emerges during brain development. Meister et al \cite{meister1991synchronous} suggested that spontaneous and spatially organized synchronized bursts prevalent in the developing retina guide the self-organization of cortical receptive fields. In this light, mathematical models of the retina and its emergent retinal waves were built \cite{godfrey2007retinal}, and analytical solutions were obtained regarding the self-organization of wiring between the retina and the LGN \cite{haussler1983development,willshaw1976patterned,eglen2009self,swindale1996development,swindale1980model}. These models have been essential for understanding how self-organization functions in the brain, but haven't been generalized to growing complex architectures that can compute. One of the most successful attempts at growing a 3D model of neural tissue from simple precursor units was demonstrated by Zubler et. al \cite{zubler2009framework} that defined a set of minimal rules that could result in the growth of morphologically diverse neurons. Although their networks were grown from single units, they weren't functional as they weren't equipped to perform any task. To bridge this gap, in this paper we attempt to grow and self-organize functional neural networks from a single precursor unit.

\section{Bio-inspired developmental algorithm} \label{mathematical-model}

In our procedure, the pooling architecture emerges through two processes, growth of a layered neural network followed by self-organization of its inter-layer connections to form defined `pools' or receptive fields. As the protocol for growing a network is relatively straightforward, our emphasis in the next few sections is on the self-organization process, following which we will combine the growth of a layered neural network with its self-organization in the penultimate section of this paper. 

We, first, abstract the natural development strategy as a mathematical model around a set of input sensor nodes in the first layer (similar to retinal ganglion cells) and processing units in the second layer (similar to cells in the LGN). 

Self-organization comprises of two major elements: (1) A \textbf{spatiotemporal wave generator} in the first layer driven by noisy interactions between input-sensor nodes and (2) A \textbf{local learning rule} implemented by units in the second layer to learn the ``underlying'' pattern of activity generated in the first layer. These two elements are inspired by mechanisms deployed by the early visual system, which spontaneously triggers retinal waves that tile the light-insensitive retina, that further serve as signals to wire the retina to higher visual areas in the brain \cite{wong1999retinal, anton2019prenatal}.

\subsection{Spontaneous spatiotemporal wave generator}

The first layer can serve as a noise-driven spatiotemporal wave generator when (1) its constituent sensor-nodes are modeled via an appropriate dynamical system and (2) when these nodes are connected in a suitable topology. In this paper, we model each sensor node using the classic Izikhevich neuron model \cite{izhikevich2003simple} (dynamical system model), while the input layer topology is that of local-excitation and global-inhibition, a motif that is ubiquitous across various biological systems \cite{kutscher2004local, xiong2010cells}. A minimal dynamical systems model coupled with the local-excitation and global-inhibition motif has been analytically examined in the supplemental materials to demonstrate that these key ingredients are \textit{sufficient} to serve as a spatiotemporal wave generator.
\begin{figure}[H]
    \centering
    \includegraphics[width=0.9\linewidth]{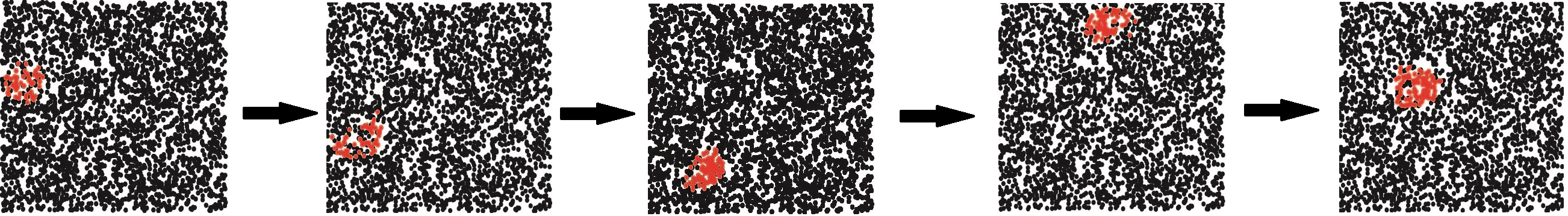}
   \caption{\textbf{Emergent spatiotemporal waves tile the first layer.} The red-nodes indicate active-nodes (firing), black nodes refer to silent nodes and the arrows denote the direction of time.}
   \label{fig:spatiotemporal_wave(a)}
   \vspace{-3mm}
\end{figure}
The \textbf{Izhikevich model} captures the activity of every sensor node ($v_i(t)$) through time, noisy behavior of individual nodes (through $\eta_i(t)$) and accounts for interactions between nodes within the same layer defined by a synaptic adjacency matrix ($S_{i,j}$). These equations are elaborated in Box-1. The \textbf{input layer topology} (local excitation, global inhibition) is defined by the synaptic adjacency matrix ($S_{i,j}$). Every node in the first layer makes excitatory connections with nodes within a defined local excitation radius. $S_{i,j}$ = 5, when distance between nodes $i$ and $j$ are within the defined excitation radius of 2 units; $d_{ij}$ $\leq$ 2. Each node has decaying inhibitory connections with other nodes present above a defined global inhibition radius ($S_{i,j}$ = -2 exp(-$d_{ij}$/10), when distance between nodes $i$ and $j$ are above a defined inhibition radius of 4 units; $d_{ij}$ $\geq$ 4) (see supporting information).  

On implementing a model of the resulting dynamical system, we observe the emergence of spontaneous spatiotemporal waves that tile the first layer for specific parameter regimes (see figure \ref{fig:spatiotemporal_wave(a)} and videos in supplemental materials). 

\begin{tcolorbox} \label{box:izhikevich}
\textbf{Dynamical model for input-sensor nodes in the lower layer (layer-I):}
\begin{equation}
    \frac{\mathrm{d}v_i}{\mathrm{d}t}= 0.04v_i^2 + 5v_i + 140 - u_i + \sum_{j=1}^N S_{i,j}\mathcal{H}(v_j-30) + \eta_i(t) \\ 
\end{equation}

\begin{equation}
    \frac{\mathrm{d}u_i}{\mathrm{d}t} = a_i(b_iv_i - u_i) 
\end{equation}

with the auxiliary after-spike reset:
\[
    v_i(t) > 30, \mathrm{ then:}
\begin{cases}
    v_i(t+\Delta t) = c_i \\
    u_i(t+\Delta t) = u_i(t) + d_i
\end{cases}
\]

where: ($1$) $v_i$ is the activity of sensor node $i$; ($2$) $u_i$ captures the recovery of sensor node $i$; ($3$) $S_{i,j}$ is the connection weight between sensor-nodes $i$ and $j$; ($4$) $N$ is the number of sensor-nodes in layer-I; ($5$) Parameters $a_i$ and $b_i$ are set to 0.02 and 0.2 respectively, while $c_i$ and $d_i$ are sampled from the distributions $\mathcal{U}(-65,-50)$ and $\mathcal{U}(2,8)$ respectively. Once set for every node, they remain constant during the process of self-organization. The initial values for $v_i(0)$ and $u_i(0)$ are set to -65 and -13 respectively for all nodes. These values are taken from Izhikevich's neuron model \cite{izhikevich2003simple}; ($6$) $\eta_i(t)$ models the noisy behavior of every node $i$ in the system, where $<\eta_i(t)\eta_j(t')>$ = $\sigma^2$ $\mathrm{\delta_{i,j}}\mathrm{\delta(t-t')}$. Here, $\mathrm{\delta_{i,j}}$, $\mathrm{\delta(t-t')}$ are Kronecker-delta and Dirac-delta functions respectively, and $\sigma^2$ = 9; ($7$) $\mathcal{H}$ is the unit step function:
\[
    \mathcal{H}(v_i-30) = 
\begin{cases}
    1,& v_i\geq30 \\
    0,& v_i<30. \\
\end{cases}
\]

\end{tcolorbox}

\subsection{Local learning rule} \label{sec:localLearningRule}

Having constructed a spontaneous spatiotemporal wave generator in layer-I, we  implement a local learning rule in layer-II that can learn the activity wave pattern in the first layer and modify its inter-layer connections to generate a pooling architecture.  Many neuron inspired learning rules  can learn a sparse code from a set of input examples \cite{olshausen1996emergence}. Here, we model processing units as rectified linear units (ReLU) and implement a modified Hebbian rule for tuning the inter-layer weights to achieve the same. Individual ReLU units compete with one another in a winner take all fashion.  

\begin{figure}[H]
    \centering
    \includegraphics[width=0.7\linewidth]{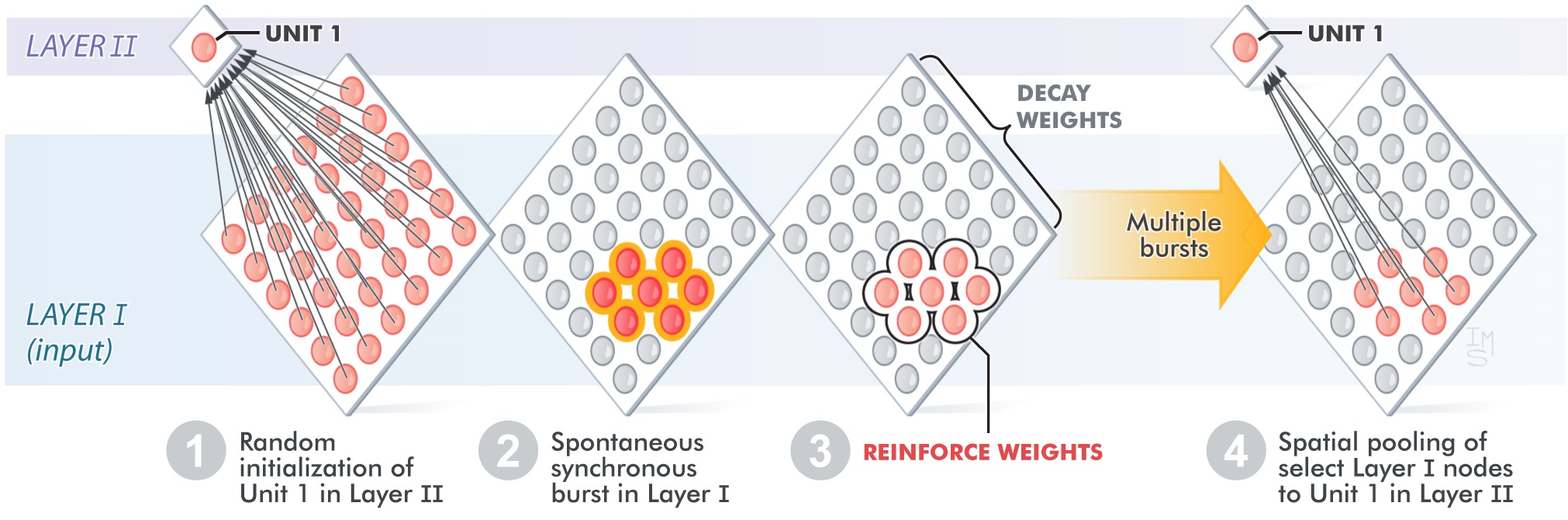}
    \caption{Learning rule}
    \label{fig:learningRule}
    \vspace{-3mm}
\end{figure}

Initially, every processing unit in the second layer is connected to all input-sensor nodes in the first layer. As the emergent activity wave tiles the first layer, at most a single processing unit in the second layer is activated due to the winner-take-all competition. The weights connecting the activated unit in the second layer to the input-sensor nodes in the first layer are updated by the modified Hebbian rule (Box-2). Weights connecting active input-sensor nodes and activated processing units are reinforced while weights connecting inactive input-sensor nodes and activated processing units decay (cells that fire together, wire together). Inter-layer weights are updated continuously throughout the self-organization process, ultimately resulting in the pooling architecture (See figure-\ref{fig:learningRule} and supplemental materials). 

\begin{tcolorbox}
\textbf{Modifying inter-layer weights} 

\[
    w_{i,j}(t+1) = 
\begin{cases}
    w_{i,j}(t) + \eta_{learn}\mathcal{H}(v_i(t)-30)y_{j}(t+1) & \text{$y_{j}(t+1)$ $>$ 0}\\
    w_{i,j}(t) & \text{otherwise} \\
\end{cases}
\]

where: ($1$) $w _{i,j}(t)$ is the weight of connection between sensor-node $i$ and processing unit $j$ at time `t' (inter-layer connection); ($2$) $\eta_{learn}$ is the learning rate; ($3$) $\mathcal{H}(v_i(t)-30)$ is the activity of sensor node $i$ at time `t'; and ($4$) $y_{j}(t)$ is the activation of processing unit $j$ at time `t'. \\

Once all the weights $w_{i,j}(t+1)$ have been evaluated for a processing unit $j$, they are mean-normalized to prevent a weight blow-up. This ensures that the mean strength of weights for processing unit $j$ remains constant during the self-organization process.

\end{tcolorbox}

Having coupled the spontaneous spatiotemporal wave generator and the local learning rule, we observe that an initially fully connected two-layer network (figure-4a) becomes a pooling architecture, wherein input-sensor nodes that are in close proximity to each other in the first layer have a very high probability of connecting to the same processing unit in the second layer (figure-4b $\&$ 4c). More than $95\%$ of the sensor-nodes in layer-I connect to processing units in layer-II (higher layer) through well-defined pools, ensuring that spatial patches of nodes connected to units in layer-II tile the input layer (figure-4d). Tiling the input layer ensures that most sensor nodes have an established means of sending information to higher layers after the self-organization of the pooling layer.

\begin{figure}[H]
    \centering
    \includegraphics[width=0.95\linewidth]{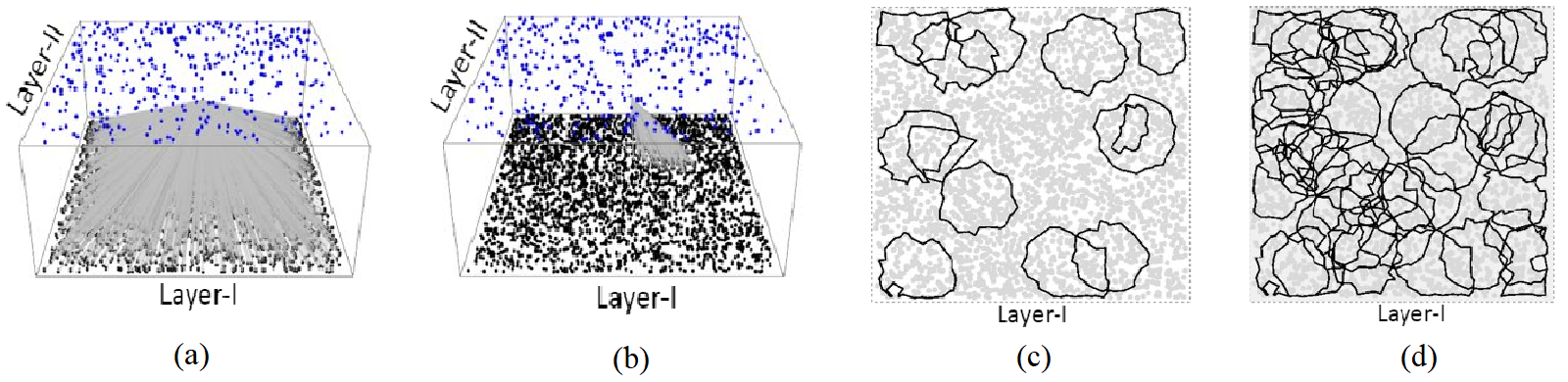}
    \caption{\textbf{Self-organization of Pooling layers.} (a) The initial configuration, wherein all nodes in the lower layer are connected to every unit in the higher layer. (b) After the self-organization process, a pooling architecture emerges, wherein every unit in layer-II is connected to a spatial patch of nodes in layer-I. (a,b) Here, connections from nodes in layer-I to a single unit in layer-II (higher layer) are shown. 
    (c) Each contour represents a spatial patch of nodes in layer-I connected to a single unit in layer-II.
    (d) More than 95$\%$ of the nodes in layer-I are connected to units in the layer-II through well-defined pools, as the spatial patches tile layer-I completely.}
    \label{fig:selfOrganize_pool}
    \vspace{-3mm}
\end{figure}

\section{Features of the developmental algorithm}

In this section, we show that spatiotemporal waves can emerge and travel over layers with arbitrary geometries and even in the presence of defective sensor-nodes. Since activity waves can form independent of macroscopic features of the input layer, our algorithm can construct pools over sensor layers with curved or irregular geometries and also in the presence of defects or holes. As the local structure of sensor-node connectivity (local excitation and global inhibition) in the input layer is conserved over a broad range of macroscale geometries (Figure-5a), we observe a traveling activity wave in input layers with arbitrary geometries, which when coupled to a learning rule in layer-II forms a pooling architecture (refer to supplemental information for an analytical treatment). 
Furthermore, we demonstrate that the size and shape of the emergent spatiotemporal wave can be tuned by altering the topology of sensor-nodes in the layer. Coupling the emergent wave in layer-I with a learning rule in layer-II leads to localized receptive fields that tile the input layer.

Together, the wave and the learning rule endow the developmental algorithm with useful properties:
(i) \textbf{Flexibility}: Spatial patches of sensor-nodes connected to units in layer-II can be established over arbitrary input-layer geometries. In Figure-5a, we show that an emergent spatiotemporal wave on a ring-shaped input layer coupled with the local learning rule (section-3.2) in layer-II, results in a pooling architecture. Flexibility to form pooling layers on arbitrary input-layer geometries is useful for processing data acquired from unconventional sensors, like charge-coupled devices that mimic the retina \cite{sandini1993retina}. (ii) \textbf{Robustness}: Spatial patches of sensor-nodes connected to units in layer-II can be established in the presence of defective sensor nodes in layer-I. As shown in figure-5b, we initially self-organize a pooling architecture for a fully functioning set of sensor-nodes in the input-layer. To test robustness, we ablate a few sensor-nodes in the input-layer (captioned 'DN'). Following this perturbation, we observe that the pooling architecture re-emerges, wherein spatial-pools of sensor-nodes, barring the damaged ones, re-form and connect to units in layer-II. (iii) \textbf{Reconfigurable}: The size and shape of spatial pools generated can be modulated by tuning the structure of the emergent traveling  wave (figure-5c $\&$ 5d). In figure-5e, we show that the size of spatial-pools can be altered in a controlled manner by modifying the topology of layer-I nodes. Wave-$x$ in the legend corresponds to an emergent wave generated in layer-I when every node in layer-I makes excitatory connections to other nodes in its 2 unit radius and inhibitory connections to every node above $x$ unit radius. This topological change alters the properties of the emergent wave, subsequently changing the resultant spatial-pool size. The histograms corresponding to these legends capture the distribution of spatial-pool sizes over all pools generated by a given wave-$x$. The histogram also highlights that the size of emergent spatial-pools are tightly regulated for every wave-configuration.

\begin{figure}[H]
    \centering
    \includegraphics[width=0.8\linewidth]{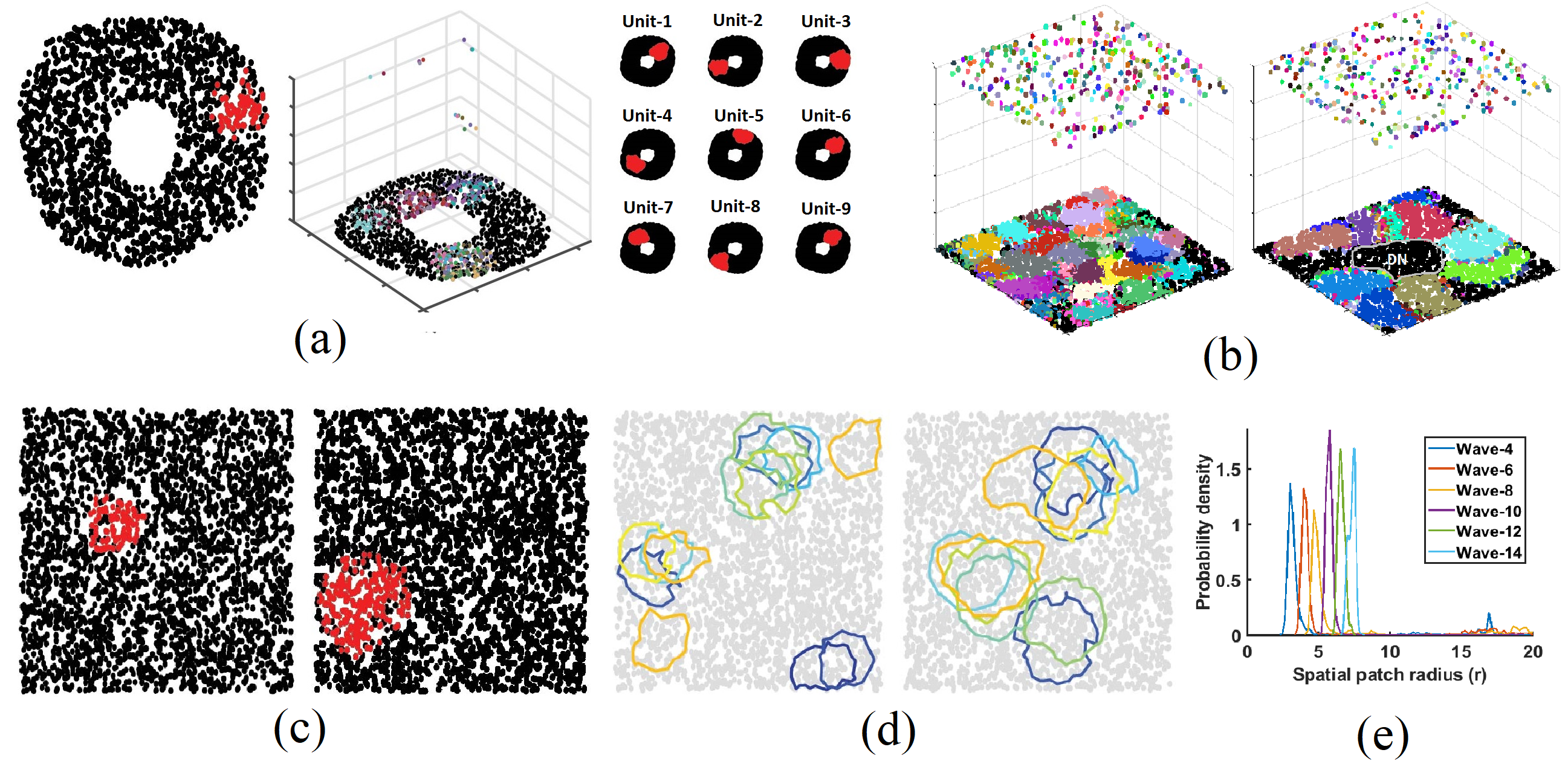}
    \caption{\textbf{Features of the developmental algorithm.}  (a) \textbf{Self-organization of pooling layers for arbitrary input-layer geometry.} (a) The left most image is a snapshot of the traveling wave as it traverses layer-I; Layer-I has sensor-nodes arranged in an annulus geometry; red nodes refer to firing nodes. On coupling the spatiotemporal wave in layer-I to a learning rule in layer-II, a pooling architecture emerges. The central image refers to the 3d visualization of the pooling architecture, while each subplot in the right-most image depicts the spatial patch of nodes in layer-I connected to a single processing unit in layer-II. (b) \textbf{Self-organization of pooling layers are robust to input layer defects} (b) The figure on the left depicts a self-organized pooling layer when all input nodes are functioning. Once these inter-layer connections are established, a small subset of nodes are damaged to assess if the pooling architecture can robustly re-form. The set of nodes within the grey boundary, titled `DN', are defective nodes. The figure on the right corresponds to pooling layers that have adapted to the defects in the input layer, hence not receiving any input from the defective nodes.(c,d,e) \textbf{Pooling layers are reconfigurable.} (c) By altering layer-I topology (excitation/inhibition radii), we can tune the size of the emergent spatial wave. The size of the wave is 6 A.U (left) and 10 A.U (right). (d) Altering the size of the emergent spatial wave tunes the emergent pooling architecture. The size of the pools obtained are 4 A.U (left), obtained from a wave-size of 6 A.U and a pool-size of 7 A.U (right), obtained from a wave-size of 10 A.U. (e) A large set of spatial-pools are generated for every size-configuration of the emergent wave. The distribution of spatial-pool sizes over all pools generated by a specific wave-size are captured by a kernel-smoothed histogram. Wave-4 in the legend corresponds to a histogram of pool-sizes generated by an emergent wave of size 4 A.U (blue line). We observe that spatial patches that emerge for every configuration of the wave have a tightly regulated size.}
    \label{fig:tune_pool}
    \vspace{-3mm}
\end{figure}


\section{Growing a neural network}

As the developmental algorithm is flexible to varying scaffold geometries and tolerant to malfunctioning nodes, it can be implemented for growing a system, enabling us to push AI in the direction towards being more 'life-like' by reducing human involvement in the design of complex functioning architectures. The growth paradigm implemented in this section has been inspired by mechanisms that regulate neocortical development \cite{rakic2000radial,bystron2008development}.



The process of growing a layered neural network involves two major sub-processes. One, every `node' can divide horizontally to produce daughter nodes that populates the same layer; Two, every node can divide vertically to produce daughter processing units that migrate upwards to populate higher layers. Division is stochastic and is controlled by a set of random variables. Having defined the 3D scaffold, we seed a single unit (figure-6a). As horizontal and vertical division ensues to form the layered neural network, inter-layer connections are modified based on the emergent activity wave in layer-I and a learning rule (section-3.2) in layer-II, to form a pooling architecture. A detailed description of the growth rule-set coupled with a flow chart governing the growth of the network is appended to the supplemental materials.



\begin{figure}[h]
    \centering
    \subfloat[\label{fig:dev_1}]{\includegraphics[width=.2\textwidth]{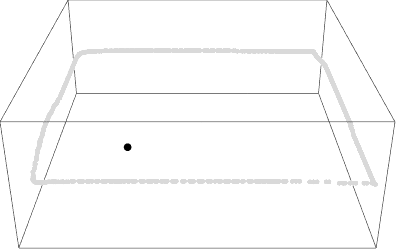}}\hfill
    \subfloat[\label{fig:dev_2}]{\includegraphics[width=.2\textwidth]{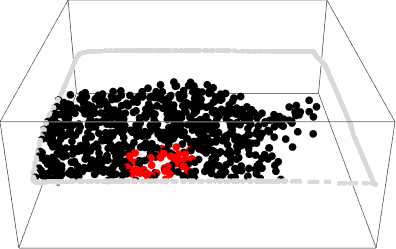}}
    \qquad
    \subfloat[\label{fig:dev_3}]{\includegraphics[width=.2\textwidth]{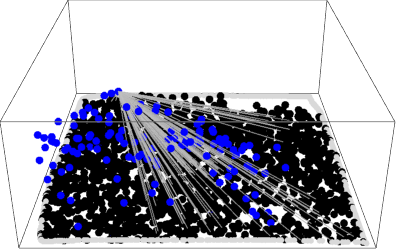}}\hfill
    \subfloat[\label{fig:dev_4}]{\includegraphics[width=.2\textwidth]{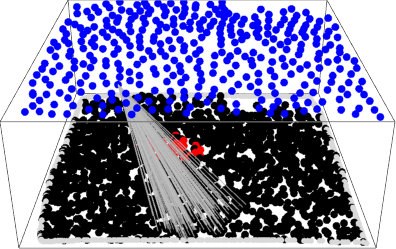}}\hfill
    \caption{\textbf{Growing a layered neural network} (a) A single computational "cell" (black node) is seeded in a scaffold defined by the grey boundary. (b) Once this "cell" divides, daughter cells make local-excitatory and global-inhibitory connections. As the division process continues, noisy interactions between nodes results in emergent spatiotemporal waves (red-nodes). (c) Some nodes within layer-I divide to produce daughter cells that migrate upwards to form processing units (blue nodes). The connections between the two layers are captured by the lines that connect a single unit in a higher layer to nodes in the first layer (Only connections from a single unit are shown).(d) After a long duration, the system reaches a steady state, where two layers have been created with an emergent pooling architecture.}
    \vspace{-3mm}
    \label{fig:dev_neuralSystem}
\end{figure}
    
Having intertwined the growth of the system and self-organization of inter-layer connections, we make the following interesting observations: (1) spatiotemporal waves emerge in the first layer much before the entire layer is populated (figure-\ref{fig:dev_2}), (2) self-organization of inter-layer connections commences before the layered network is fully constructed (figure-\ref{fig:dev_3}) and (3)  Over time, the system reaches a steady state as the number of `cells' in the layered network remains constant and most processing units in the second layer connect to a pool of nodes in the first layer, resulting in the pooling architecture (figure-\ref{fig:dev_4}). Videos of networks growing on arbitrary scaffolds are added to the supplemental materials.


\section{Growing functional neural networks}

In the previous section, we demonstrated that we can successfully grow multi-layered pooling networks from a single unit. In this section, we show that these networks are functional.

We demonstrate functionality of networks grown and self-organized from a single unit (figure-7c) by evaluating their train and test accuracy on a classification task. Here, we train networks to classify images of handwritten digits obtained from the MNIST dataset (figure-7e). To interpret the results, we compare it with the train/test accuracy of hand-crafted pooling networks and random networks. Hand-crafted pooling networks have a user-defined pool size for all units in layer-II (figure-7b), while random networks have units in layer-II that connect to a random set of nodes in layer-I without any spatial bias (figure-7d), effectively not forming a pooling layer. 

To test functionality of these networks, we couple the two-layered network with a linear classifier that is trained to classify hand-written digits from MNIST on the basis of the representation provided by these three architectures (hand-crafted, self-organized and random networks). 
We observe that self-organized networks classify with a 90$\%$ test accuracy, are statistically similar to hand-crafted pooling networks (90.5$\%$, p-value = 0.1591) and are statistically better than random networks (88$\%$, p-value = 5.6 x 10$^{-5}$) (figure-7a). This performance is consistent over multiple self-organized networks. These results show that self-organized neural networks are functional and can be adapted to perform conventional machine-learning tasks, with the big add-on of being autonomously grown from a single unit. 

\begin{figure}[t]
    \centering
    \includegraphics[width=0.95\linewidth]{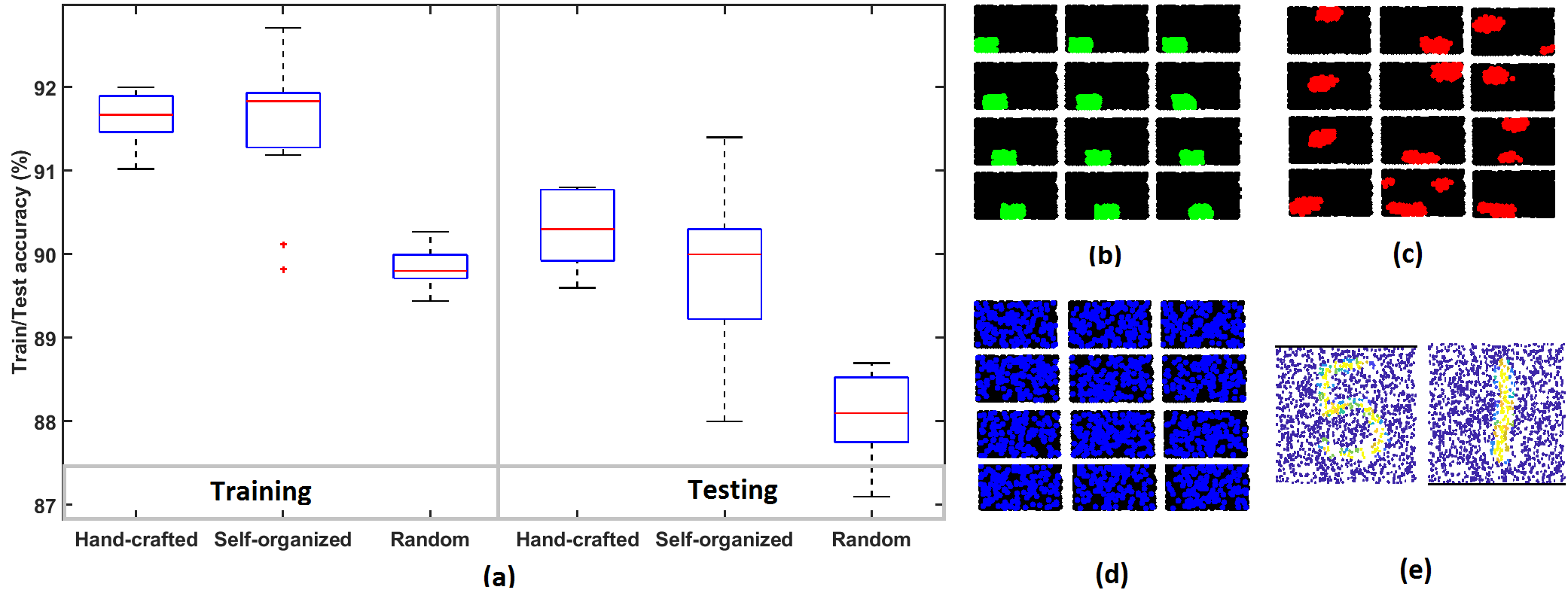}
    \caption{\textbf{Networks grown from a single unit are functional.}  
    Three kinds of networks are trained and tested on images obtained from the MNIST database. We use 10000 training samples and 1000 testing samples. The 3 kinds of networks are: (i) Hand-crafted, (ii) Self-organized networks and (iii) random networks. This procedure is run over n=11 networks to ensure that the developmental algorithm always produces functional networks. (a) The box-plot captures the training and testing accuracy of these 3 networks. We notice that the testing accuracy of self-organized networks is comparable to that of to that of hand-crafted networks (p-value = 0.1591$>$0.05) and are much better than random networks (p-value = 5.6 x 10$^{-5}$). (b,c,d) Each unit in the second layer is connected to a set of nodes in the lower layer. The set it is connected to are defined by the green, red or blue nodes in the subplots shown. (b) Hand-crafted (c) Self-organized and (d) Random-basis.(e) Two MNIST images as seen in the first layer.  }    
    \label{fig:func_network}
    \vspace{-3mm}
\end{figure}

\section{Discussion}

In this paper, we address a pertinent question of how artificial computational machines could be built autonomously with limited human intervention. Currently, architectures of most artificial systems are obtained through heuristics and hours of painstaking parameter tweaking. Inspired by the development of the brain, we have implemented a developmental algorithm that enables the robust growth and self-organization of functional layered neural networks. 

Implementation of this framework brought many crucial questions concerning neural development to our attention. Neural development is classically defined by discrete steps, one proceeding the other. However this isn't the case, as development is a continuous flow of events with multiple intertwined processes \cite{arlotta2017editorial}. Our work on growing artificial systems got us interested in how critical times of different developmental processes are controlled, and whether they were controlled by an internal clock.

The work also reinforces the significance of brain-inspired mechanisms for initializing functional architecture to achieve generalization for multiple tasks. A peculiar instance in the animal kingdom would be the presence of precocial species, animals whose young ones are functional immediately after they are born. One mechanism that enables functionality immediately after birth is spontaneous activity that assists in maturing neural circuits much before the animal receives any sensory input. Although we have shown how a layered architecture (mini-cortex) can emerge through spontaneous activity in this paper, our future work will focus on growing multiple components of the brain, namely a hippocampus and a cerebellum, followed by wiring these regions in a manner useful for an organism's functioning. This paradigm of growing mini-brains in-silico will allow us to (i) explore how different components in a biological brain interact with one another and guide our design of neuroscience experiments and (ii) equip us with systems that can autonomously grow, function and interact with the environment in a more `life-like' manner. 

\subsubsection*{Acknowledgments}

We would like to thank Markus Meister, Carlos Lois, Erik Winfree, Naama Barkai for their invaluable contribution for shaping the early stages of the work. We also thank Alex Farhang, Jerry Wang, Tony Zhang, Matt Rosenberg, David Brown, Ben Hosheit, Varun Wadia, Gautam Goel, Adrianne Zhong and Nasim Rahaman for their constructive feedback and key edits that have helped shape this paper.


\printbibliography[heading=subbibliography]

\end{refsection}

\newpage 

\section*{\centering \huge Supplementary Material}


\setcounter{section}{0}
\setcounter{equation}{0}
\setcounter{figure}{0}
\setcounter{table}{0}
\setcounter{page}{1}
\renewcommand{\theequation}{S\arabic{equation}}
\renewcommand{\thefigure}{S\arabic{figure}}
\renewcommand{\thesection}{S\arabic{section}}

\begin{refsection}
\section{Mathematical model}

\subsection{Dynamical model for input sensor nodes}

Input sensor nodes are modeled using the Izhikevich neuron model. This is used primarily because it has the least number of parameters for accurately modeling neuron-like activity and the parameter regimes that produce different neuronal firing states have been well characterized earlier \cite{izhikevich2003simple}.

\begin{tcolorbox} \label{suppbox:izhikevich}
\textbf{Dynamical model for input-sensor nodes in the lower layer (layer-I):}
\begin{equation}
    \frac{\mathrm{d}v_i}{\mathrm{d}t}= 0.04v_i^2 + 5v_i + 140 - u_i + \sum_{j=1}^N S_{i,j}\mathcal{H}(v_j-30) + \eta_i(t) \\ 
\end{equation}

\begin{equation}
    \frac{\mathrm{d}u_i}{\mathrm{d}t} = a_i(b_iv_i - u_i) 
\end{equation}

with the auxiliary after-spike reset:
\[
    v_i(t) > 30, \mathrm{ then:}
\begin{cases}
    v_i(t+\Delta t) = c_i \\
    u_i(t+\Delta t) = u_i(t) + d_i
\end{cases}
\]

where: ($1$) $v_i$ is the activity of sensor node $i$; ($2$) $u_i$ captures the recovery of sensor node $i$; ($3$) $S_{i,j}$ is the connection weight between sensor-nodes $i$ and $j$; ($4$) $N$ is the number of sensor-nodes in layer-I; ($5$) Parameters $a_i$ and $b_i$ are set to 0.02 and 0.2 respectively, while $c_i$ and $d_i$ are sampled from the distributions $\mathcal{U}(-65,-50)$ and $\mathcal{U}(2,8)$ respectively. Once set for every node, they remain constant during the process of self-organization. The initial values for $v_i(0)$ and $u_i(0)$ are set to -65 and -13 respectively for all nodes. These values are taken from Izhikevich's neuron model \cite{izhikevich2003simple}; ($6$) $\eta_i(t)$ models the noisy behavior of every node $i$ in the system, where $<\eta_i(t)\eta_j(t')>$ = $\sigma^2$ $\mathrm{\delta_{i,j}}\mathrm{\delta(t-t')}$. Here, $\mathrm{\delta_{i,j}}$, $\mathrm{\delta(t-t')}$ are Kronecker-delta and Dirac-delta functions respectively, and $\sigma^2$ = 9; ($7$) $\mathcal{H}$ is the unit step function:
\[
    \mathcal{H}(v_i-30) = 
\begin{cases}
    1,& v_i\geq30 \\
    0,& v_i<30. \\
\end{cases}
\]

\end{tcolorbox}

\subsection{Topology of input-sensor nodes} \label{app:topology_input}

\begin{figure}[H]
    \centering
   \subfloat[\label{fig:topology_input}]{{\includegraphics[width=0.4\linewidth]{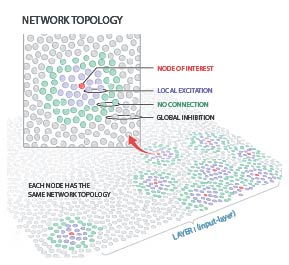}}}\hfill
    \subfloat[\label{fig:wtsTrend}]{{\includegraphics[width=0.4\linewidth]{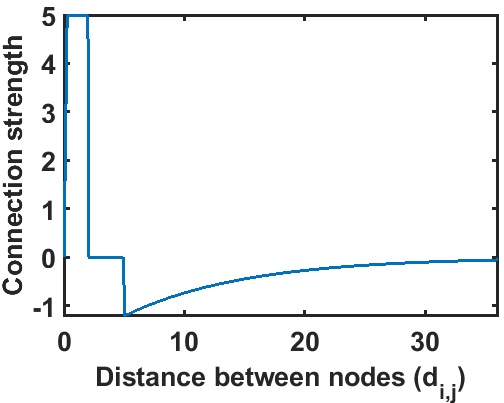}}}
    \caption{Topology of sensor-node connections: Every node is connected to other nodes in the layer within a radius $r_{e}$ via a positive weight, not connected to nodes positioned at a distance between $r_{e}$ and $r_{i}$ and connected to nodes at a distance larger than $r_{i}$ with a decaying negative weight.}
    \vspace{-3mm}
\end{figure}

The nodes in the lower layer (layer-I) are arranged in a local-excitation, global inhibition topology, with a ring of nodes that have neither excitation or inhibition (zero weights) between the excitation and inhibition regions. We have observed that this ring of no connections between the excitation and inhibition regions gives us a good handle over the emergent wave size. This is detailed in Box-\ref{box:topology} and depicted in figure-\ref{fig:topology_input}.

\begin{tcolorbox} \label{box:topology}
\textbf{Topology of input-sensor nodes in layer-I:}\\ \\
This topology is pictorially depicted in figure-\ref{fig:topology_input} and mathematically defined below:
\[
    S_{i,j} = 
\begin{cases}
    l,& d_{i,i} \leq r_{e}\\
    m \exp(\frac{-d_{i,j}}{10}), & d_{i,j} \geq r_{i} \\
    0 & r_{e} <d_{i,j}< r_{i}
\end{cases}
\]
where:
\begin{itemize}
    \item $S_{i,j}$ is the connection weight between sensor-nodes $i$ and $j$ 
    \item $d_{i,j}$ is the Euclidean distance between sensor-nodes $i$ and $j$ in layer-I
    \item $r_e$ is the local excitation radius ($r_e$ = 2)
    \item $r_i$ is the global inhibition radius (all nodes present outside this radius are inhibited) ($r_i$ = 4)
    \item $l$ is the magnitude of excitation ($l$ = 5)
    \item $m$ is the magnitude of inhibition ($m$ = -2)
\end{itemize}
\end{tcolorbox}

\subsection{Modeling Processing units and winner-take-all strategy} \label{app:process_units}

Processing units are modeled as Rectified linear units (ReLU) associated with an arbitrary threshold. Although the threshold is randomly initialized, it is updated during the process of self-organization. Its update depends entirely on the activity trace of the processing unit it is associated with. We also require that at every time point, at most a single processing unit in layer-II be activated by the emergent patterned activity in layer-I. To enforce this, we let the processing units, modeled as ReLU units compete with each other in a winner-take-all (WTA) manner. This ensures that at every time point, at most a single unit in layer-II responds to the patterned activity in the input layer.

Each processing unit in layer-II is modeled by the equation given below:
\begin{equation}
    y_{j}(t) = \mathcal{W} [\text{max} (0, \sum_{i=1}^{N} w_{i,j}(t) \mathcal{H}(v_i(t)-30))]
\end{equation}

Here, the $\text{max}(0,x)$ is the implementation of a rectified linear unit (ReLU); $\mathcal{H}(v_i(t)-30)$ is the threshold activity of sensor node $i$ (in layer-I) at time `t';  $y_{j}(t)$ is the activation of processing unit $j$ (in layer-II) at time `t'; $w_{i,j}^t$ is the connection weight between sensor-node $i$ and processing unit $j$ at time `t'; $N$ is the number of sensor-nodes in layer-I and $\mathcal{W}$ refers to the winner-take-all mechanism that ensures a single winning processing unit. 

The winner-take-all function implemented in layer-II is mathematically elaborated below:
\[
    \mathcal{W}[y_{j}(t)] = 
\begin{cases}
    \mathrm{max}(0, y_{j}(t)-c_j(t)), & \text{if } y_{j}(t) > y_{k}(t) \hspace{2mm}  \forall k \in [1,...j-1,j+1,...,M] \\
    0 & \text{otherwise} \\
\end{cases}
\]

Here, $y_{j}(t)$ is the activation of processing unit $j$ (in layer-II) at time `t'; $c_j(t)$ is the threshold for processing unit $j$ at time `t' and $M$ is the number of processing units in layer-II. 
Every processing unit is modeled as a ReLU with an associated threshold ($c_j$). Although this threshold is arbitrarily initialized, they are updated during the process of self-organization. The update depends on the number of times the connections between processing units and nodes in layer-I are updated, and it's described below. 

To implement this, we keep track of the number of times connections between a specific processing unit and sensor nodes in layer-I are updated over the course of 1000 time-points. $z_j(t)$ captures the number of times connections between processing unit-j and sensor-nodes in layer-I are updated. 

\textbf{Keeping track of the synaptic changes per processing unit:}

\[
    z_{j}(t+1) = 
\begin{cases}
    z_{j}(t) + 1 & \text{if }(y_{j}(t)>0) \\
    0 & \text{if ($t$ mod 1000) = 0} \\
    z_j(t) & \text{otherwise}
    
\end{cases}
\]

The threshold for a processing unit is updated based on the number of connections that were altered in the past 1000 time points between that processing unit and sensor-nodes in layer-I.

\textbf{Updating the threshold for every processing unit:} \\
\[
    c_j(t+1) = 
\begin{cases}
    \mathrm{max}(y_{j}(t),y_{j}(t-1),...,y_{j}(0))/5, & \parbox{4cm}{\text{if ($t$ mod 1000) = 0 }AND $z_{j}(t)$ $<$ 200} \\
    c_j(t) & \text{otherwise} \\
\end{cases}
\]

Here, $w _{i,j}(t)$ is the weight of connection between sensor-node $i$ and processing unit $j$ at time `t'; $\eta_{learn}$ is the learning rate; $y_{j}^t$ is the activation of processing unit $j$ at time `t'; $z_j(t)$ is the number of synaptic modifications made to unit $j$ until time `t'; ($t$ mod 1000) is the remainder when $t$ is divided by 1000 and $c_j(t)$ is the activation threshold for processing unit $j$ at time `t'.

The emergent wave in layer-I coupled with the learning rule implemented by processing units in layer-II are sufficient to self-organize pooling architectures.

\section{Growing a neural network} \label{app:growNN}

We demonstrate that by defining a minimal set of `rules' for a single computational `cell', we can grow a layered network, followed by the  self-organization of its inter-layer connections to form pooling layers.



In order to grow a layered network, we define a 3D scaffold as well as seed the first layer in the scaffold with a computational `cell' (figure-\ref{fig:dev_1}). The major attributes of nodes in the first layer are: 
\begin{itemize}
    \item $v_i(t)$ : activity of node $i$ modeled by the Izhikevich equation \cite{izhikevich2003simple}
    \item $clockH_i$ : records the age of the `cell', allowing horizontal division (division within the same layer) until it reaches a certain age
    \item $HFlim_i$ : the maximum divisions permitted for node $i$
    \item $VCD_i$ : a binary variable that records whether node $i$ has vertically divided or not. Vertical division is the process when a `cell' divides and its daughter `cells' migrate upwards to form processing units that populate higher layers.
\end{itemize}


\begin{figure}[h]
    \centering
    \subfloat[\label{fig:dev_supp1}]{\includegraphics[width=.2\textwidth]{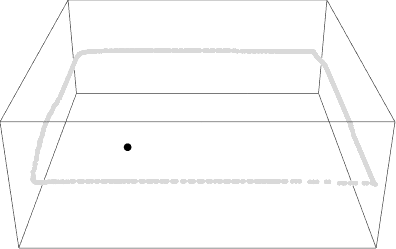}}\hfill
    \subfloat[\label{fig:dev_supp2}]{\includegraphics[width=.2\textwidth]{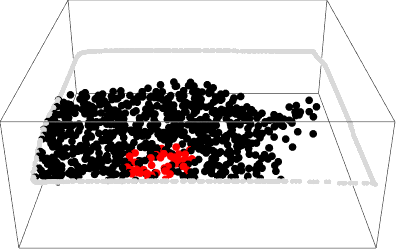}}
    \qquad
    \subfloat[\label{fig:dev_supp3}]{\includegraphics[width=.2\textwidth]{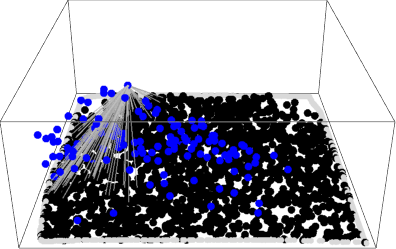}}\hfill
    \subfloat[\label{fig:dev_supp4}]{\includegraphics[width=.2\textwidth]{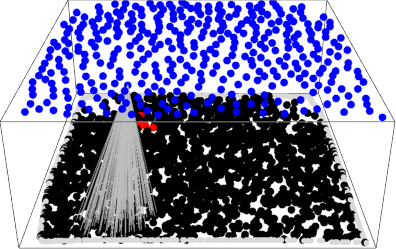}}\hfill
    \caption{\textbf{Growing a layered neural network} (a) A single computational "cell" (black node) is seeded in a scaffold defined by the grey boundary. (b) Once this "cell" divides, daughter cells make local-excitatory and global-inhibitory connections. As the division process continues, noisy interactions between nodes results in emergent spatiotemporal waves (red-nodes). (c) Some nodes within layer-I divide to produce daughter cells that migrate upwards to form processing units (blue nodes). The connections between the two layers are captured by the lines that connect a single unit in a higher layer to nodes in the first layer (Only connections from a single unit are shown).(d) After a long duration, the system reaches a steady state, where two layers have been created with an emergent pooling architecture.}
    \vspace{-3mm}
    \label{fig:dev_SuppneuralSystem}
\end{figure}


\subsection{User-defined Growth Parameters}

\begin{table}[H]
\centering
\begin{tabular}{|l|l|l|}
\hline
\textbf{Parameter}  & \textbf{Value} & \textbf{Description} \\ \hline
HCD\_AGE            & 25            & The maximum time a cell can pursue horizontal division\\ \hline
HF\_MAX             & 40             & The maximum number of divisions a single cell can pursue\\ \hline
R\_HDIV             & 1             & Critical radius I\\ \hline
R\_VDIV             & 1             & Critical radius II\\ \hline
THRESH\_HDIV        & 3               & The maximum number of cells permitted within a radius (R\_HDIV) \\ \hline
\end{tabular}
\end{table}

\subsection{Growth Process}

\textbf{Step: 1}: \\ \\
A single computational `cell' endowed with the following attributes is seeded on a 3D scaffold. The attributes and values that a seeded computational `cell' is endowed with is mentioned in the table below. The first column indicates attributes, second column denotes the initial values that they take and the third column is a description of the attribute. 

\begin{table}[H]
\centering
\begin{tabular}{|l|l|l|}
\hline
\textbf{Cell attribute}  & \textbf{Initialization} & \textbf{Description} \\ \hline
v$_i$            & -65            & Initialize activity of node $i$\\ \hline
clockH$_i$             & 0        & Initializing clock to 0, for every newly divided daughter cell\\ \hline
HFlim$_i$            & HF\_MAX    & Initializing the max divisions to HF$\_$MAX for the seeded cell. \\ \hline
VCD$_i$            & 0            & Before vertical division, VCD$_i$ = 0; After vertical division, VCD$_i$ = 1; \\ \hline
\end{tabular}
\end{table}

\subsubsection{Step: t $\rightarrow$ t+1}

A random cell $i$ is sampled from the input layer. \\ \\
If the cell hasn't crossed the critical age threshold (clockH$_i$ $<$ HCD$\_$AGE) and the number of cells within a radius (R$\_$HDIV) is below the density threshold (numCells$_i$(R$\_$HDIV) $<$ THRESH$\_$HDIV), the cell divides horizontally to form daughter cells that populate the same layer. The clockH is reset to zero for the daughter cells, however the HFlim attribute of the daughter cells is one less than their parent to keep track of the number of divisions.  \\ \\
If it hasn't reached the critical age threshold, but has a local density above the defined density threshold, it remains quiescent and a new `cell' is sampled. \\ \\  
A cell $i$ can divide vertically only if the cell has reached the critical age threshold (clockH$_i$ = HCD$\_$AGE) and cells in its local vicinity (with radius :- R$\_$VDIV) haven't divided vertically. As mentioned in an earlier section, a binary variable VCD$_i$ keeps track of whether a cell has divided vertically or not. 

When a cell divides vertically, one daughter cell occupies the parent's position on layer-I, while the other daughter cell migrates upwards. The daughter cell that migrates upwards initially makes a single connection with its twin on layer-I, which gets modified with time, resulting in a pool of nodes in layer-I making connections with a single unit in the higher layer (pooling architecture). 

\subsubsection{Termination condition}

The local rules that control horizontal division and vertical division are active throughout and prevent the system from blowing up, with respect to the number of nodes in each layer. It has been observed that the system reaches a steady state, as the number of `cells' in both layers remain constant. 

\begin{figure}[H]
    \centering
    \includegraphics[width=1.05\linewidth]{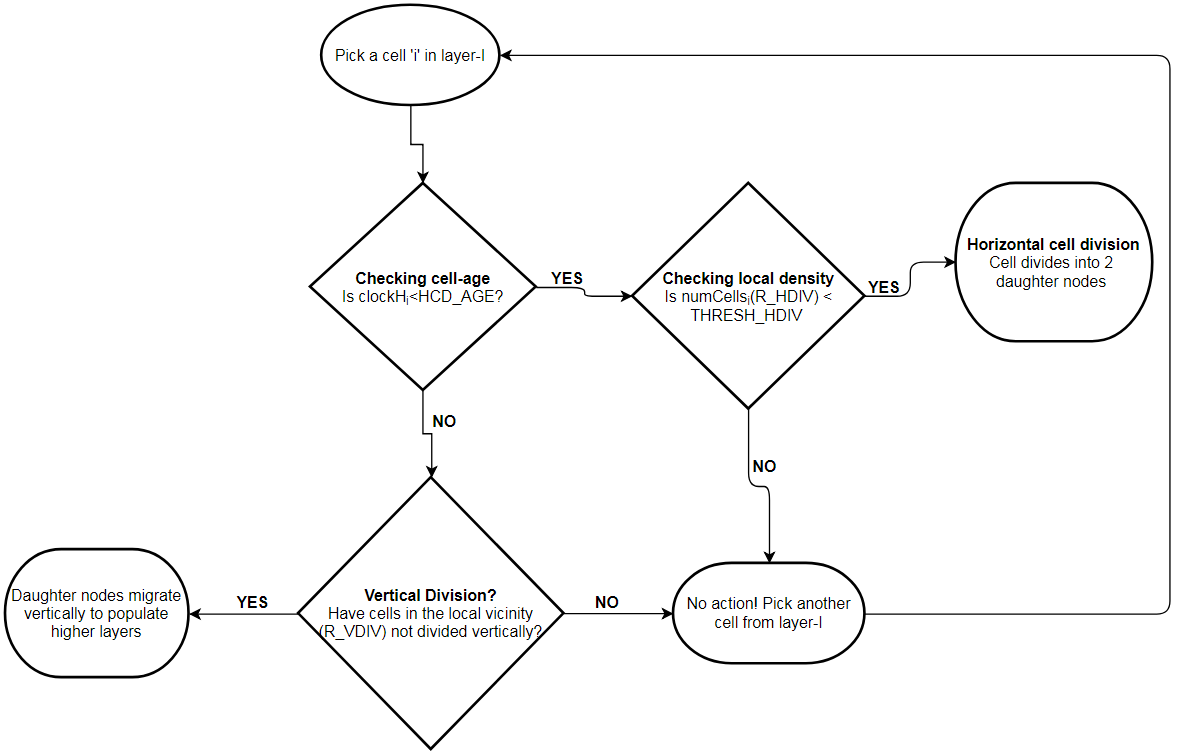}
    \caption{Growth flowchart}
    \label{fig:growth_flowchart}
\end{figure}

\subsection{Growing neural networks on arbitrary scaffolds (Results)}

Videos of multi-layered networks growing on arbitrary scaffolds can be viewed by visiting this link: [\url{https://drive.google.com/open?id=1YtFEvWHTU9HWl760V8lEr9Heapx0sUdh}]

\section{Minimal model for observing emergent spatiotemporal waves}

In this section, we provide an analytical solution for the emergence of a spatiotemporal wave through noisy interactions between constituent nodes in the same layer.

As we stated in the main-text, the key ingredients for having a layer of nodes function as a spatiotemporal wave generator are:
\begin{itemize}
    \item Each sensor-node should be modeled as a dynamical systems model
    \item Sensor-nodes should be connected in a suitable topology (here, local excitation ($r_e < 2$ and global inhibition ($r_i > 4$).
\end{itemize}

On modeling all nodes in the system using a simple set of ODE's, we highlight the conditions required for observing a stationary bump in a network of spiking sensor-nodes and to observe instability of the stationary bump resulting in a traveling wave. 

\subsection{Arranging sensor-nodes in a line}

\begin{figure}[H]
    \centering
    \includegraphics[width=0.5\linewidth]{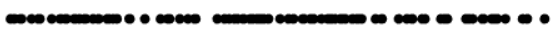}
    \caption{Sensor nodes arranged in a line}
    \label{fig:lineArrangement}
\end{figure}

We choose a configuration where $N$ sensor-nodes are randomly arranged in a line (as shown in figure-\ref{fig:lineArrangement}).

The activity of $N$ sensor nodes, arranged in a line as in figure-\ref{fig:lineArrangement}, are modeled using a minimal ODE model as described below: 
\begin{equation}
    \begin{split}
        \tau_{d}\frac{\mathrm{d}x(u_i,t)}{\mathrm{d}t} = -x(u_i,t) + \sum_{u_j \in \mathcal{U}} S(u_i,u_j)\mathcal{F}(x(u_j,t))\hspace{3mm} \forall i \in {1,...,N}
    \end{split}
\end{equation}

Here, $u_i$ represents the position of nodes on a line; $x(u_i,t)$ defines the activity of sensor node positioned at $u_i$ at time $t$; $S_{u_i,u_j}$ is the strength of connection between nodes positioned at $u_i$ and $u_j$; $\tau_{d}$ controls the rate of decay of activity; $\mathcal{U}$ is the set of all sensor nodes in the system ($u_1$,$u_2$,...,$u_N$) for $N$ sensor nodes; and $\mathcal{F}$ is the non-linear function required to convert activity of nodes to spiking activity. Here, $\mathcal{F}$ is the heaviside function with a step transition at 0.

\begin{figure}[H]
    \centering
    \includegraphics[width=0.4\linewidth]{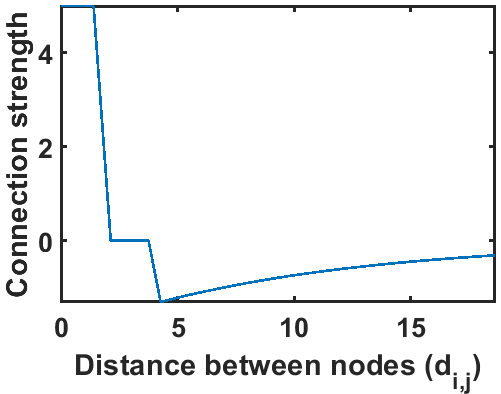}
    \caption{strength of connections between sensor-nodes}
    \label{fig:topology_sensors}
\end{figure}

Each sensor-node has the same topology of connections, ie fixed strength of positive connections between nodes within a radius $r_e$, no connections from a radius $r_{e}$ to $r_{i}$, and decaying inhibition above a radius $r_{i}$. This is depicted in figure-\ref{fig:topology_sensors}

\subsubsection{Fixed point analysis}

We determine the stable activity states of nodes placed in a line by a fixed point analysis, similar to what Amari developed in \cite{amari1977dynamics} for the case when there are infinite nodes.

\begin{equation}
\begin{split}
    x(u_i) &= \sum_{u_j \in \mathcal{U}} S(u_i,u_j)\mathcal{F}(x(u_j))\hspace{3mm} \forall i \in {1,...,N}
\end{split}
\end{equation}


On solving this system of non-linear equations simultaneously, we get a fixed point ie a vector $x^{*}$ $\in$ $\mathcal{R}^N$, corresponding to the activity of $N$ sensor nodes positioned at ($u_1$,$u_2$,...,$u_N$). To assess their spiking from the activity of sensor-nodes, we have 
\begin{equation}
    s_i = \mathcal{F}(x(u_i)) \hspace{3mm} \forall i \in {1,...,N}
\end{equation}

\begin{figure}
    \centering
   \subfloat[\label{fig:1d_1bump}]{{\includegraphics[width=0.3\linewidth]{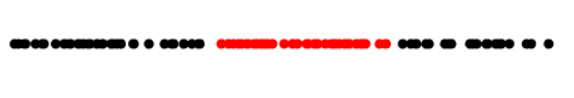}}}\hfill
    \subfloat[\label{fig:1d_1bump2}]{{\includegraphics[width=0.3\linewidth]{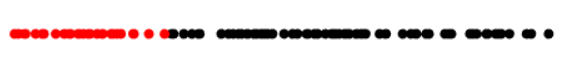}}}
    \qquad
    \subfloat[\label{fig:1d_2bumps}]{{\includegraphics[width=.3\linewidth]{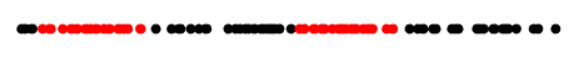}}}\hfill
    \caption{\textbf{Fixed points:} Multiple fixed points are obtained by solving $N$ non-linear equations simultaneously. Some of the solutions obtained are: (a) a single bump at the center, (b) a single bump at one of the edges and (c) two bumps of activity. }
    \vspace{-3mm}
\end{figure}

As the weight matrix ($S_{u_i,u_j}$) used incorporates the local excitation ($r_e < 2$) and global inhibition ($r_i > 4$) ( figure-\ref{fig:topology_sensors}), we get solutions with a single bump of activity (figure-\ref{fig:1d_1bump}), two bumps of activity (figure-\ref{fig:1d_2bumps}) or a state when all nodes are active. 

\subsubsection{Stability of fixed points}

To assess the stability of these fixed points, we evaluate the eigenvalues of the Jacobian for this system of differential equations. As there are $N$ differential equations, the Jacobian ($\mathbb{J}$) is an $N$x$N$ matrix. 

\begin{equation}
    \begin{split}
        \frac{\mathrm{d}x(u_i,t)}{\mathrm{d}t} &= \frac{-x(u_i,t)}{\tau_{d}} + \sum_{u_j \in \mathcal{U}} \frac{S(u_i,u_j)\mathcal{F}(x(u_j))}{\tau_d} \\
        \frac{\mathrm{d}x(u_i,t)}{\mathrm{d}t} &= f_i(u_1, u_2, ..., u_N) \\
        f_i(u_1, u_2, ..., u_N) &= \frac{-x(u_i)}{\tau_d} + \sum_{u_j \in \mathcal{U}} \frac{S(u_i,u_j)\mathcal{F}(x(u_j))}{\tau_d} \\
        \mathbb{J}(i,j) &= \frac{\partial f_i(u_1, u_2, ..., u_N)}{\partial x(u_j)}
    \end{split}
\end{equation}

On evaluating the Jacobian ($\mathbb{J}$) at the fixed points obtained (x$^*$), we get: 
\begin{equation}
    \begin{split}
        \mathbb{J}(i,i) &= \frac{\partial f_i}{\partial x(u_i)} \\
        \mathbb{J}(i,i) &= \frac{-1}{\tau_d} \\
        \mathbb{J}(i,j) &= S(u_i,u_j)\mathcal{F}^{'}(x(u_j))\frac{\partial x(u_j)}{x(u_j)} \\
        \mathbb{J}(i,j) &= S(u_i,u_j)\delta(x(u_j)) \\
        \mathbb{J}(i,j) &= 0 \hspace{2mm} \forall x(u_j) \neq 0
    \end{split}
\end{equation}

Here, $\mathcal{F}$ is the Heaviside function and its derivative is the dirac-delta($\delta$); where, $\delta(x)$ = 0, for $x \neq 0$ and $\delta(x)$ = $\infty$ for $x = 0$. 

For a fixed point, where $x^{*}(u_k) \neq 0$, $\forall k \in$ ${1,...,N}$, the Jacobian is a diagonal matrix with $\frac{-1}{\tau_d}$ in its diagonals. This implies that the eigenvalues of the Jacobian are $\frac{-1}{\tau_d}$ ($\tau_d > $0), which assures that the fixed point $x* \in \mathcal{R}^N$ is a stable fixed point. 

\subsubsection{Destabilizing the fixed point}

With the addition of high amplitude of gaussian noise to the ODE's described earlier, we can effectively destabilize the fixed point, resulting in a traveling wave. The equations with the addition of a noise term are:  

\begin{equation}
     \tau_{d}\frac{\mathrm{d}x(u_i,t)}{\mathrm{d}t} = -x(u_i,t) + \sum_{u_j \in \mathcal{U}} S(u_i,u_j)\mathcal{F}(x(u_j,t)) + \eta_i(t) \hspace{3mm} \forall i \in {1,...,N}
\end{equation}
Here, $\eta_i(t)$ models the noisy behavior of every node $i$ in the system, where $<\eta_i(t)\eta_j(t')>$ = $\sigma^2$ $\mathrm{\delta_{i,j}}\mathrm{\delta(t-t')}$. Here, $\mathrm{\delta_{i,j}}$, $\mathrm{\delta(t-t')}$ are Kronecker-delta and Dirac-delta functions respectively, and $\sigma^2$ captures the magnitude of noise added to the system. 

The network of sensor nodes is robust to a small amplitude of noise ($\sigma^2 \in$ (0,4)), while a larger amplitude of noise ($\sigma^2$>5) can destabilize the bump, forcing the system to transition to another bump in its local vicinity.  Continuous addition of high amplitudes of noise forces the bump to move around in the form of traveling waves. The behavior is consistent with the linear stability analysis because noise can push the dynamical system beyond the envelop of stability for a given fixed point solution. 

\subsection{Arranging sensor nodes in a 2D square}

\begin{figure}[H]
    \centering
    \includegraphics[width=0.2\linewidth]{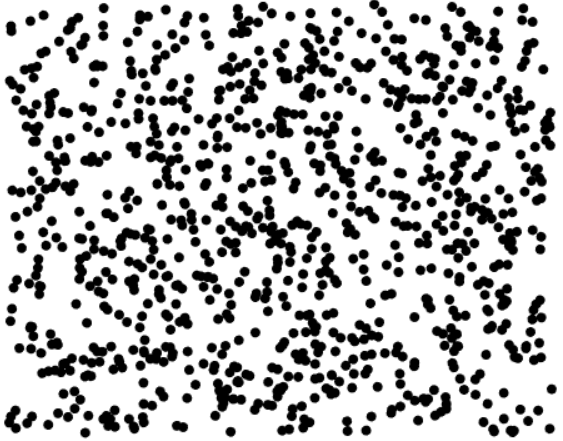}
    \caption{Sensor nodes placed arbitrarily on a square plane}
    \label{fig:2DsquareArr}
    \vspace{-3mm}
\end{figure}

In this section, we arrange $N$ sensor nodes arbitrarily on a 2-dimensional square as shown in figure-\ref{fig:2DsquareArr}, with the same local structure (local excitation and global inhibition).

The activity of these sensor nodes are modeled using the minimal ODE model described earlier (in equation-4).

We obtain the fixed points ({x$^{*}$ $\in \mathcal{R}^N$}), by solving $N$ simultaneous non-linear equations using BBsolve \cite{varadhan2009bb}. We notice that the fixed point solutions have a variable number of activity bumps in the 2D plane as shown in figure-8a,8b $\&$ 8c.

\begin{figure}[H]
    \centering
   \subfloat[\label{fig:2d_1bump}]{{\includegraphics[width=0.25\linewidth]{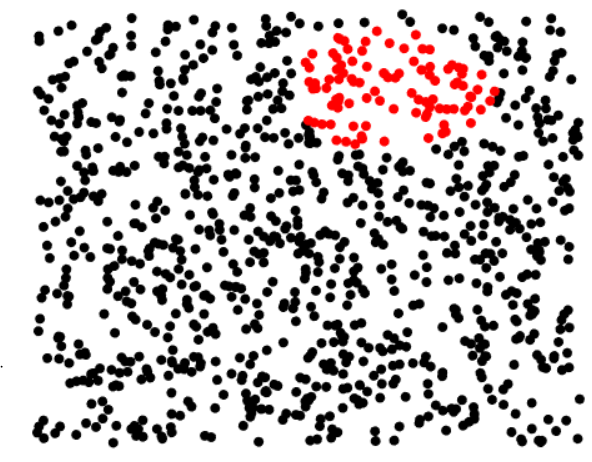}}}\hfill
    \subfloat[\label{fig:2d_2bump}]{{\includegraphics[width=0.25\linewidth]{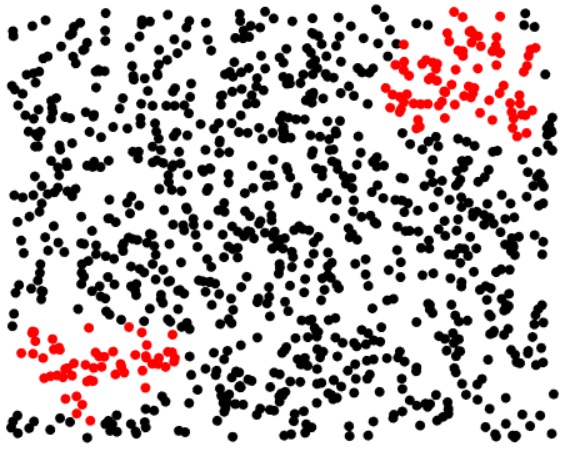}}}\hfill
    \qquad
    \subfloat[\label{fig:2d_3bump}]{{\includegraphics[width=.25\linewidth]{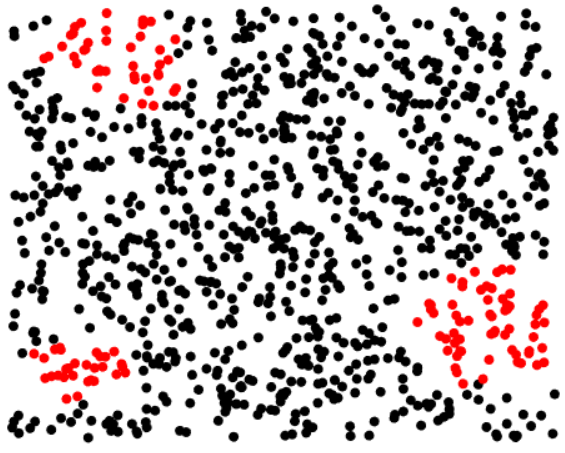}}}
    \caption{\textbf{Stable Fixed points:} Multiple fixed points are obtained by solving $N$ non-linear equations simultaneously. Some of the solutions obtained are: (a) a single bump, (b) two bumps and (c) three bumps of activity. }
    \vspace{-3mm}
\end{figure}

\subsection{Arranging sensor nodes on a 2D sheet of arbitrary geometry}

In this section, we arrange sensor nodes on a 2D sheet in any arbitrary geometry as shown in figure 9. Although the macroscopic geometry of the sheet changes, the local structure of sensor nodes in conserved (ie local excitation and global inhibition).

The fixed points are evaluated by simultaneously solving the non-linear system of equations. We notice that the bumps are stable fixed points even when sensor nodes are placed on a 2-dim sheet of arbitrary geometry. 

\begin{figure}[H]
    \centering
   \subfloat[\label{fig:arbitGeo1}]{{\includegraphics[width=0.2\linewidth]{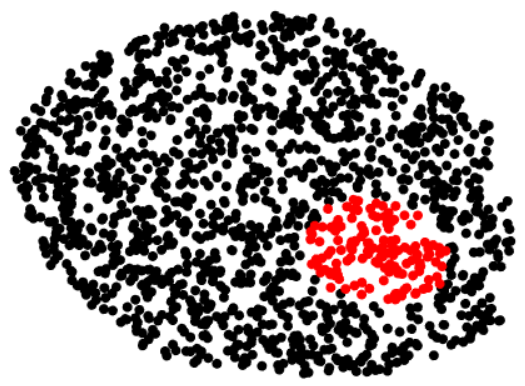}}}\hfill
    \subfloat[\label{fig:arbitGeo2}]{{\includegraphics[width=0.2\linewidth]{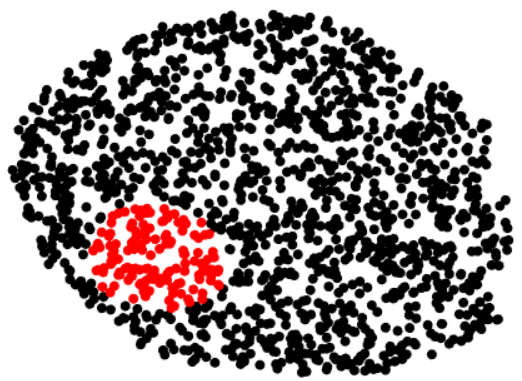}}}\hfill
    \qquad
    \subfloat[\label{fig:arbitGeo3}]{{\includegraphics[width=.2\linewidth]{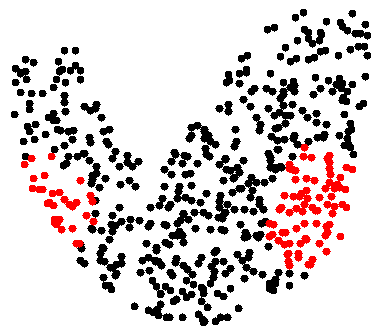}}}\hfill
    \qquad
    \subfloat[\label{fig:arbitGeo4}]{{\includegraphics[width=.2\linewidth]{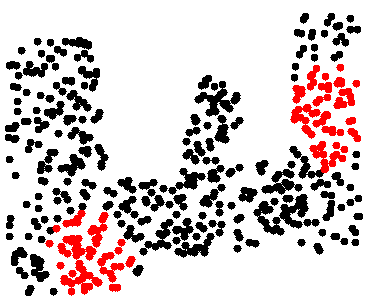}}}
    \caption{\textbf{Stable Fixed points:} Multiple fixed points are obtained by solving $N$ non-linear equations simultaneously. Some of the solutions obtained are: (a,b) a single bump for a circular geometry (c,d) two bumps of activity for arbitrary geometry }
    \vspace{-3mm}
\end{figure}

\section{Growing functional neural networks}

We estimate functionality of networks grown and self-organized from a single unit by evaluating their train and test accuracy on a classification task. Here, we train networks to classify images of handwritten digits obtained from the MNIST dataset. To interpret the results, we compare it with the train/test accuracy of hand-crafted pooling networks and random networks. Hand-crafted pooling networks have a user-defined pool size for all units in layer-II, while random networks have units in layer-II that connect to a random set of nodes in layer-I without any spatial bias, effectively not forming a pooling layer. 

To test functionality of these networks, we couple the two-layered network with a linear classifier that is trained to classify hand-written digits from MNIST on the basis of the representation provided by these three architectures (hand-crafted, self-organized and random networks). 

The first two layers in the network serve as feature extractors, while the last layer behaves like a perceptron. The optimal classifier is learnt by minimizing the least square error between the output of the network and a desired target. However, there isn't any back-propagation through the entire network. In essence, the architecture grown through the developmental algorithm remains fixed, performing the task of latent feature representation, while the classifier learns how to match these latent features with a set of task-based labels. 

\subsection{Setting up the pooling architecture}

The first two layers of the network correspond to the pooling architecture grown by the developmental algorithm. The input is fed to the first layer, while the units in the second layer, that are connected to spatial pools in layer-I, extract features from these inputs. 

Let $x \in \mathcal{R}^N$ be the input data (for $N$ sensor nodes) and the weights connecting the first and second layer be $W_1 \in \mathcal{R}^{M \times N}$ (for $M$ processing units). The features extracted in layer-II are: $y = \mathcal{F}(W_1x)$. Here, $\mathcal{F}$ is any non-linear function applied to the transformation in order to map all the values in layer-II within the range [-1,1].

\subsection{Appending a fully connected layer}

The pooling architecture sends its feature map through a fully connected layer with $L$ nodes, with the weights connecting the set of processing units and the fully connected layer being randomly initialized as $W_2 \in \mathcal{R}^{L \times M}$. The features extracted by the fully connected layer are: $y_{FC} = \mathcal{F}(Wy)$.
$\mathcal{F}$ is the same as the one used in section-4.1.

\subsection{Classification accuracy}

The final set of weights connecting the fully connected layer to the 10 element vector (as there are 10 digit classes in the MNIST dataset) is denoted by $W_3 \in \mathcal{R}^{10 \times L}$. The output generated by the network is $y_O = W_3y_{FC}$. Let us denote the target output as $y_T$.

As we want to minimize the least square error between the target output ($y_T$) and output of the network ($y_O$), conventionally, we can perform a gradient descent. However, as it is a linear classifier, we have a closed form solution for the weight matrix ($W_3$).

\begin{equation*}
    \begin{split}
        y_O &= W_3 y_{FC}    \\
        y_T &= W_3 y_{FC} \hspace{10mm} \text{for zero error, $y_0 = y_T$} \\
        y_T y_{FC}^T &= W_3 y_{FC} y_{FC}^T \\
        W_3 &= y_T y_{FC}^T (y_{FC} y_{FC}^T) 
    \end{split}
\end{equation*}

Setting the weights between the fully connected layer and the output layer ($W_3 = y_T y_{FC}^T (y_{FC} y_{FC}^T$), we evaluate the train and test accuracy for 3 kinds of networks. (Hand-crafted pooling, self-organized and random networks). These networks differ primarily in how their first two layers are connected. The hand-programmed pooling networks are those that have a fixed size of spatial pool that connects to units in layer-II, while the random networks have no spatial pooling. 

The results are described in the main-paper and we observe that self-organized networks classify with a 90$\%$ test accuracy are statistically similar to hand-crafted pooling networks (90.5$\%$, p-value = 0.1591) and statistically better than random networks (88$\%$, p-value = 5.6 x 10$^{-5}$) (figure-7a). This performance is consistent over multiple self-organized networks. The train/test accuracy of self-organization networks highlights that growing networks through a brain-inspired developmental algorithm is potentially useful to building functional networks.


\section{Scalability: Determining the speed of self-organization of the pooling architecture as the size of the input-layer increases}
\label{app:theorem}


Here, we demonstrate that the pooling layers can be self-organized for very large input layers. Large layers are defined based on the number of sensor nodes in the layer. We observe that enforcing a spatial bias on the initial set of connections from units in layer-II to the nodes in the input layer, enables us to speed up the process of self-organization.

Our simulations show that the self-organization of pooling layers can be scaled up to large layers (with upto 50000 nodes) without being very expensive, as an increase in number of sensor-nodes results in multiple simultaneous waves tiling the input layer, effectively forming a pooling architecture in parallel.
\begin{figure}[H]
    \centering
    \subfloat[Input layer: 1500 nodes\label{fig:scale_pool(a)}]{{\includegraphics[width=0.3\linewidth]{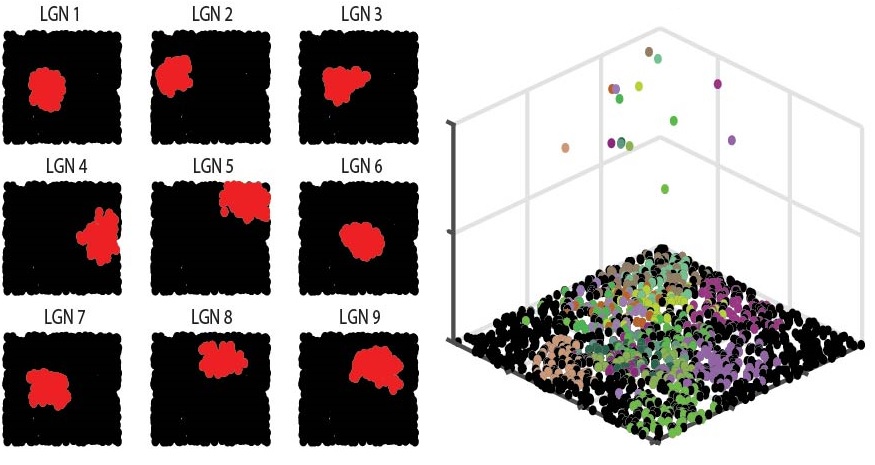}}} \hfill
    \subfloat[Input layer: 5000 nodes\label{fig:scale_pool(b)}]{{\includegraphics[width=0.3\linewidth]{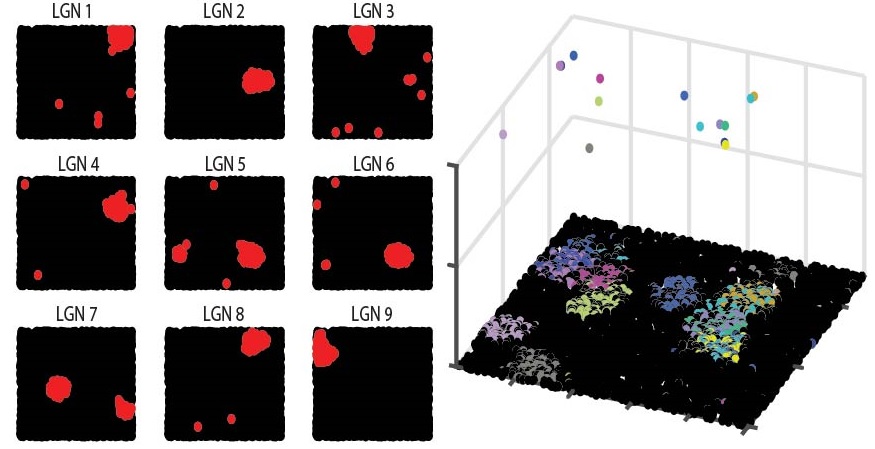}}} \hfill
    \subfloat[Input layer: 10000 nodes\label{fig:scale_pool(c)}]{{\includegraphics[width=0.3\linewidth]{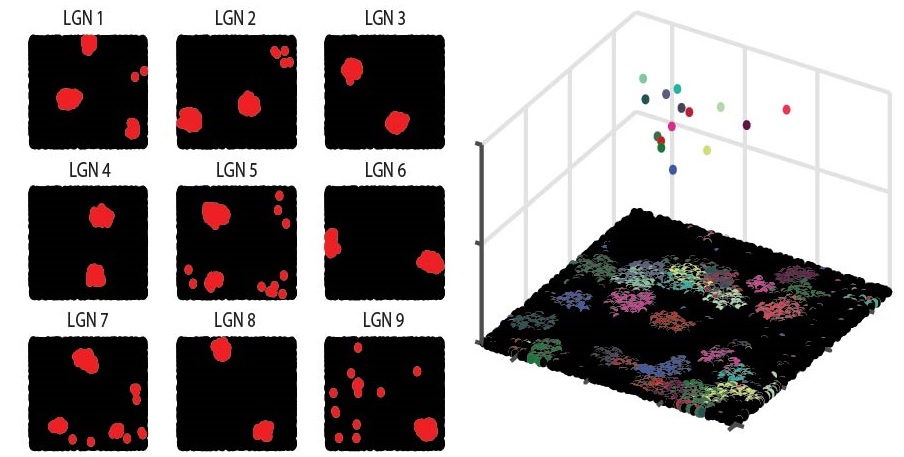}}} \hfill
     \subfloat[Time complexity for self-organization of pooling layers\label{fig:scale_pool(d)}]{{\includegraphics[width=0.5\linewidth]{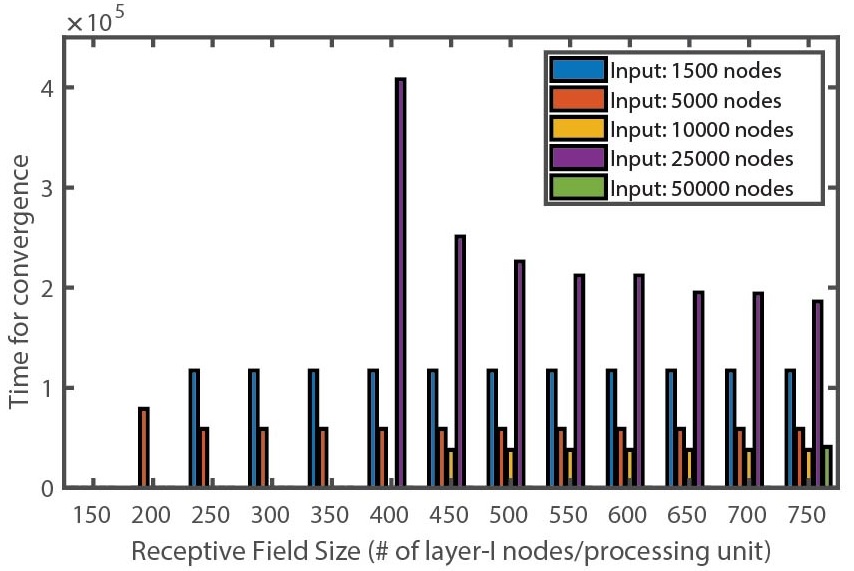}}}%
  
    \caption{\textbf{Developmental algorithm scales efficiently to very large input layers:} \small(a) Layer-I has 1500 nodes and layer-II has 400 nodes. The emergent wave in layer-I results in a single traveling wave that tiles layer-I. (b) Layer-I has 5000 nodes and layer-II has 400 nodes. The emergent wave in layer-I results in a single traveling wave that tiles layer-I. (c) Layer-I has 10000 nodes and layer-II has 400 nodes. The emergent wave in layer-I results in a multiple traveling wave that tile layer-I simultaneously. This results in a single processing unit receiving pools from different regions. (d) The histogram captures the time taken for a pooling layer to form for variable number of input sensor nodes (1500, 5000, 10000, 25000 and 50000 nodes). With an increase in the number of sensor-nodes, the speed of self-organization increases as multiple waves tile the input layer simultaneously.}
    \label{fig:scale_pool}
\end{figure}

\printbibliography[heading=subbibliography]

\end{refsection}

\end{document}